\documentclass[pdflatex,sn-nature]{sn-jnl}

\usepackage{graphicx}%
\usepackage{multirow}%
\usepackage{amsmath,amssymb,amsfonts}%
\usepackage{amsthm}%
\usepackage{mathrsfs}%
\usepackage[title]{appendix}%
\usepackage{xcolor}%
\usepackage{textcomp}%
\usepackage{manyfoot}%
\usepackage{booktabs}%
\usepackage{algorithm}%
\usepackage{algorithmicx}%
\usepackage{algpseudocode}%
\usepackage{pifont}
\usepackage{listings}%
\usepackage{tabularx}%
\usepackage{makecell}%
\usepackage{booktabs} %
\usepackage[most]{tcolorbox}

\usepackage{algorithm}
\usepackage{algpseudocode}
\usepackage{amsmath}
\usepackage{amsfonts}
\usepackage{xcolor}
\usepackage{booktabs}
\usepackage{multirow}
\usepackage{graphicx}
\usepackage{threeparttable}
\usepackage{makecell}
\usepackage[table]{xcolor}
\usepackage{pifont}
\usepackage{array}
\usepackage{booktabs}
\usepackage{hhline}

\usepackage{tikz} \usetikzlibrary{arrows.meta,positioning,fit,shapes.misc}

\newcommand{\etal}{\textit{et al.}}

\theoremstyle{thmstyleone}%
\theoremstyle{thmstyletwo}%

\theoremstyle{thmstylethree}%

\begin{document}

\title[Article Title]{Prompt Triage: Structured Optimization Enhances Vision-Language Model Performance on Medical Imaging Benchmarks}


\author*[1]{\fnm{Arnav} \sur{Singhvi}}\email{arnavs11@stanford.edu}

\author[2,3]{\fnm{Vasiliki} \sur{Bikia}}\email{bikia@stanford.edu}

\author[5]{\fnm{Asad} \sur{Aali}}\email{asadaali@stanford.edu}

\author[2,3,5]{\fnm{Akshay} \sur{Chaudhari}}\email{akshaysc@stanford.edu}

\author[2,4]{\fnm{Roxana} \sur{Daneshjou}}\email{roxanad@stanford.edu}

\affil*[1]{Stanford University, Department of Computer Science, Stanford, CA, USA}

\affil[2]{Stanford University, Department of Biomedical Data Science, Stanford, CA, USA}

\affil[3]{Stanford University, Stanford Institute for Human-Centered Artificial Intelligence, Stanford, CA, USA}

\affil[4]{Stanford University, School of Medicine, Department of Dermatology, Stanford, CA, USA}

\affil[5]{Stanford University, Department of Radiology, Stanford, CA, USA}


\abstract{Vision-language foundation models (VLMs) show promise for diverse imaging tasks but often underperform on medical benchmarks. Prior efforts to improve performance include model finetuning, which requires large domain-specific datasets and significant compute, or manual prompt engineering, which is hard to generalize and often inaccessible to medical institutions seeking to deploy these tools. 
These challenges motivate interest in approaches that draw on a model’s embedded knowledge while abstracting away dependence on human-designed prompts to enable scalable, weight-agnostic performance improvements. To explore this, we adapt the Declarative Self-improving Python (DSPy) framework for structured automated prompt optimization in medical vision-language systems through a comprehensive, formal evaluation. We implement prompting pipelines for five medical imaging tasks across radiology, gastroenterology, and dermatology, evaluating 10 open-source VLMs with four prompt optimization techniques. Optimized pipelines achieved a median relative improvement of 53\% over zero-shot prompting baselines, with the largest gains ranging from 300\% to 3,400\% on tasks where zero-shot performance was particularly low. These results highlight the substantial potential of applying automated prompt optimization to medical AI systems, particularly demonstrating significant gains for vision-based applications where accurate clinical image interpretation is crucial. 
By reducing dependence on prompt design to elict intended outputs from models, these techniques allow clinicians to focus on patient care and clinical decision-making.

Furthermore, our experiments offer scalability and preserve data privacy by demonstrating how open-source VLMs can be optimized without compute-intensive techniques that alter model weights. We publicly release our evaluation pipelines to support reproducible research and improved performance on specialized medical tasks, available at \url{https://github.com/DaneshjouLab/prompt-triage-lab}.}




\maketitle

\section{Introduction}\label{sec1}

Vision-language models (VLMs) are large-scale artificial intelligence (AI) models trained on images and paired text and can perform a range of tasks like image captioning, visual question answering (VQA), and visual reasoning. These VLM capabilities have advanced through large-scale pre-training and architectural design. Contrastive language-image pretraining (CLIP) \cite{radford2021learning} demonstrated how pre-training on hundreds of millions of image-text pairs enables strong zero-shot transfer to diverse vision tasks. Instruction-tuned and multimodal chat agents, such as MiniGPT-4 \cite{zhu2023minigpt}, LLaVA \cite{liu2023visualinstructiontuning}, and OpenFlamingo \cite{awadalla2023openflamingo}, build on general-purpose models by enabling joint reasoning over images and text. Despite these advances, state-of-the-art VLMs often underperform on specialized medical benchmarks that demand the ability to identify diagnoses or clinical concepts across a range of imaging modalities and disease states. 
To address these limitations, recent work has explored fine-tuning VLMs specifically for medical domains to improve performance. Li~\etal~\cite{li2023llava} introduced LLaVA-Med, rapidly adapted for biomedical assistance through figure-caption pairs and curriculum instruction-tuning. Extending this idea to radiology, Lin~\etal~\cite{chen2024chexagent} presented CheXagent, a vision-language foundation model trained on a large-scale chest X-ray dataset and evaluated on CheXbench, demonstrating competitive performance across eight radiological task types and improving report drafting efficiency in clinical workflows. Additionally, Lin~\etal~\cite{lin2025healthgptmedicallargevisionlanguage} proposed HealthGPT, which applies heterogeneous low-rank adaptation (H-LoRA) in a federated learning scenario to adapt a generalist VLM into a unified autoregressive model for medical visual tasks while conserving privacy across different medical sites.

Other efforts have explored prompt engineering for VLMs in medical use cases as an alternative to finetuning. Whereas finetuning adapts model parameters to new domains, prompt-based methods shape model behavior without weight updates by providing structured instructions and examples that guide reasoning and task execution. Prompting strategies such as Chain-of-Thought and self-consistency \cite{wei2022chain} have improved reasoning in LLMs and few-shot prompting approaches like those introduced by Shakeri~\etal~\cite{shakeri2024few} have demonstrated how carefully designed prompts can improve VLM performance on medical visual tasks.

Declarative Self-Improving Python (DSPy) \cite{khattab2023dspy} provides a unified framework in which prompts are expressed as modular components that can be automatically optimized within larger model prompting pipelines. Unlike aforementioned model-specific tuning approaches, DSPy enables systematic improvement of these components through declarative specifications and task-level feedback, adapting model behavior through offline optimization without modifying model weights. By leveraging metric-driven refinement at inference time, DSPy supports scalable, composable systems that generalize across models and optimization techniques, enabling robust evaluations of such techniques to assess downstream task gains.
While DSPy has shown strong potential for text-based tasks, its application to vision–language models (VLMs) and multimodal medical benchmarks remains largely unexplored.

\begin{figure}[t]
  \centering
  \includegraphics[width=\linewidth]{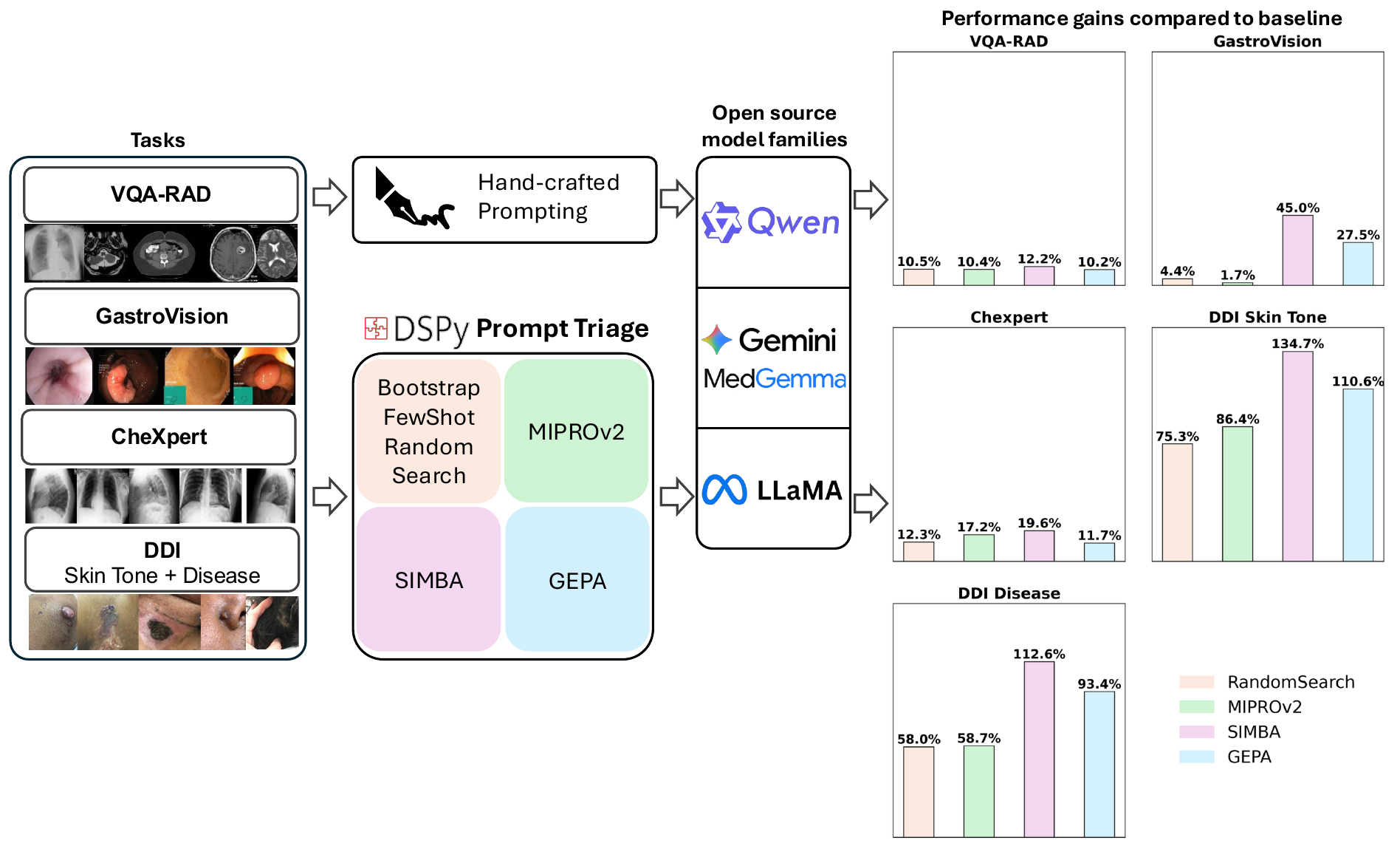}
  \caption{\textbf{Overview of DSPy Prompt Triage for medical vision-language models.} We benchmark five medical imaging tasks (VQA-RAD, GastroVision, CheXpert, DDI Skin Tone, and DDI Diagnosis) across open-source VLM families (Qwen, Gemma/MedGemma, Llama). Baselines use handwritten prompts, while DSPy Prompt Triage applies various prompt optimization techniques (BootstrapFewShotRandom Search, MIPROv2, GEPA, SIMBA), yielding performance gains over the baselines.}
  \label{fig:central_fig}
\end{figure}

In this work, we develop prompting pipelines for five challenging medical multimodal tasks and evaluate the effectiveness of prompt optimization across 10 open-source vision–language models (VLMs) of diverse providers and sizes. We build composable programs for each benchmark that themselves are agnostic to the open-source VLMs we evaluate and compare performance gains from applying prompt optimization techniques (BootstrapFewShotRandomSearch, MIPROv2, GEPA, and SIMBA) that iteratively improve each model's performance and capabilities (Figure \ref{fig:central_fig}). Notably, all improvements are achieved without training or finetuning, relying entirely on inference-time optimization.

Our study presents the first large-scale evaluation of 50 model–benchmark combinations, comparing unoptimized baselines to four prompt optimization strategies. Our experiments demonstrate median relative gains of 53\% across best-performing optimizers and extreme improvements of 300–3,400\% in low-baseline settings. In 22\% of evaluations, the best-performing optimized program for a model series’ smallest size model surpassed its family’s largest model baseline. Beyond benchmarking, we analyze optimizer effectiveness, identifying conditions and task types where prompt optimization delivers the greatest impact, and quantifying gains by model, dataset, and evaluation metric. To support reproducibility and further research, we release our evaluation pipelines and examples illustrating the extension of prompt optimization to VLMs on medical benchmarks, available at \url{https://github.com/DaneshjouLab/prompt-triage-lab}.

\section{Methods}\label{sec11}

We implement systems across four medical vision-language datasets (Table~\ref{tab:datasets}) covering five distinct image-based downstream tasks: visual question answering with a radiology concentration (VQA-RAD \cite{lau2018dataset}), multi-class endoscopy image classification (GastroVision \cite{jha2023gastrovision}), chest X-ray classification (CheXpert \cite{irvin2019chexpert}), skin tone classification (DDI - Skin Tone \cite{daneshjou2022disparities}), and dermatological diagnosis (DDI - Diagnosis \cite{daneshjou2022disparities}). Each experiment adhered to standardized train-test splits and utilized task-specific evaluation metrics to guide and assess optimization improvement over baselines: Quasi-Exact Match for VQA-RAD, F1 Score for GastroVision and CheXpert, Exact Match for DDI Skin Tone, and Top-3 Inclusion for DDI Diagnosis.

\begin{table}[ht]
\centering
\scriptsize
\renewcommand{\arraystretch}{1.2}
\begin{tabularx}{\textwidth}{l l X X c}
\toprule
\textbf{Task (Dataset)} & \textbf{Input} & \textbf{Output / Task Type} & \textbf{Evaluation Metric} & \textbf{Test Size} \\
\midrule
\makecell[l]{Radiology \\ question answering \\ (\textbf{VQA-RAD \cite{lau2018dataset}})} & \makecell[l]{Chest radiographs} & \makecell[l]{Open-ended \\ generation} & \makecell[l]{Quasi-Exact Match \\(QuasiEM)} & 451 \\
\makecell[l]{Gastroenterology \\ classification \\ (\textbf{GastroVision \cite{jha2023gastrovision}})} & Endoscopy images & \makecell[l]{Multi-class \\ disease \\ classification} & F1 score & 1,586 \\
\makecell[l]{Chest X-ray \\ classification \\ (\textbf{CheXpert \cite{irvin2019chexpert}})} & Chest X-rays & \makecell[l]{Multi-class \\ abnormality \\ classification} & F1 score & 265 \\
\makecell[l]{Skin tone \\ classification \\ (\textbf{DDI \cite{daneshjou2022disparities}})} & \makecell[l]{Dermatology \\ images} & \makecell[l]{Skin tone \\ label prediction} & \makecell[l]{Exact Match \\ (EM)} & 104 \\
\makecell[l]{Dermatology \\ diagnosis \\ (\textbf{DDI\cite{daneshjou2022disparities}})} & \makecell[l]{Dermatology \\ images} & \makecell[l]{Multi-class \\ diagnosis \\ prediction} & Top-3 inclusion rate & 104 \\
\bottomrule
\end{tabularx}
\caption{Datasets, inputs, outputs, evaluation metrics, and test set sizes used in this study.}
\label{tab:datasets}
\end{table}

For each task, we leveraged predefined prompts from a recent study evaluating general VLM performance for these tasks~\cite{jiang2024evaluating} as the baseline instruction. Input data comprises both image and text input, including the corresponding medical image, the  task objective, and optionally a list of possible choices to output for classification tasks (GastroVision, CheXpert, and both DDI tasks). In VQA-RAD, models generate open-ended responses to the corresponding questions. All prompts explicitly instruct the VLMs to perform chain-of-thought reasoning to support their final predictions.
We then evaluate prompt optimization techniques, spanning in-context learning, instruction tuning and iterative feedback-driven refinement  (BootstrapFewShotWithRandomSearch, GEPA, MIPROv2, SIMBA)~\cite{opsahl2024optimizing, agrawal2025gepareflectivepromptevolution} \footnote{https://github.com/stanfordnlp/dspy}  to systematically improve each task's baseline program:

\begin{itemize}
    \item BootstrapFewShotRandomSearch bootstraps candidate prompts by sampling small labeled example sets ("few-shots") and selects the final prompt with the highest-performing examples embedded. \textit{(Few-shot only)}
    \item MIPROv2 bootstraps possible few-shot examples, proposes diverse instruction candidates, and then uses Bayesian optimization to score and combine them to find the best combination prompt. \textit{(Few-shot + instruction tuning)}
    \item SIMBA iteratively edits instructions or examples, evaluates them on mini-batches, and uses feedback signals to guide each update towards the best combination prompt. \textit{(Few-shot + instruction tuning)}
    \item GEPA reflects on execution traces to generate feedback and uses evolutionary search to refine and preserve the strongest prompt candidates until convergence. \textit{(Instruction tuning)}
\end{itemize}

All DSPy programs and optimization configurations used in this study are publicly released in our repository (\url{https://github.com/DaneshjouLab/prompt-triage-lab}) to support reproducibility and further experimentation.

\section{Experiments and Results}\label{sec2}

We standardize evaluation by randomly sampling 500 examples from each task’s training set for prompt optimization and evaluating each optimized pipeline on the held-out test sets: VQA-RAD (451 examples), GastroVision (1,586), CheXpert (265), and DDI tasks (Skin Tone and Disease, 104 each). To assess robustness, we repeat this process across three independent bootstrap trials, each with a different set of 500 training examples driving the optimization, and report the average test set evaluation score for each optimized program over the three runs.
We benchmarked a range of open-source pre-trained vision-language models of varying scales and architectures, including: Qwen2.5-VL series (3B, 7B, 32B, 72B), Gemma series (gemma3-4B, medgemma-4b-it [a medically fine-tuned Gemma variant], gemma3-12B, gemma3-27B), and Llama3.2 series (llama3.2-11B, llama3.2-90B). 
We leave model parameters and weights unchanged, only evaluating the optimized pipelines with prompt changes reflected during model inference on the test set.

Evaluation metrics are respective to the task type:  Quasi-Exact Match (QuasiEM) for VQA-RAD, F1 Scores for GastroVision and CheXpert tasks, Exact Match (EM) for DDI Skin Tone classification, and Top-3 Inclusion rate for DDI Diagnosis.




\newcommand{\meanCI}[2]{#1{\scriptsize$\;\pm\;$#2}}

\begin{table}[ht]
\centering
\scriptsize
\resizebox{\textwidth}{!}{%
\begin{tabular}{ll|c|c|c|c|c!{\vrule width 1.2pt}}
\hhline{======!{\vrule width 1.2pt}=}
\textbf{Model} & \makecell{\textbf{Optimizer}\\\scriptsize{Metric}} & 
\makecell{\textbf{VQA-RAD}\\\scriptsize{QuasiEM}} &
\makecell{\textbf{GastroVision}\\\scriptsize{F1Score}} &
\makecell{\textbf{Chexpert}\\\scriptsize{F1Score}} &
\makecell{\textbf{DDI Skin Tone}\\\scriptsize{EM}} &
\makecell{\textbf{DDI Diagnosis}\\\scriptsize{Top3 Incl.}} \\
\hhline{------!{\vrule width 1.2pt}-}

\multirow{5}{*}{\texttt{Qwen2.5-VL-3B}} 
  & Baseline     & 29.5 & 7.8 & 16.9 & 16.3 & 8.7 \\
  & RandomSearch & \meanCI{41.0}{8.2} & \meanCI{17.5}{26.2} & \meanCI{15.6}{14.1} & \meanCI{50.6}{40.4} & \meanCI{20.2}{4.8} \\
  & MIPROv2      & \meanCI{43.9}{6.0} & \meanCI{14.3}{28.9} & \meanCI{20.0}{1.2} & \meanCI{52.9}{36.0} & \cellcolor{yellow}\textbf{\meanCI{26.3}{15.2}} \\
  & SIMBA        & \cellcolor{yellow}\textbf{\meanCI{45.1}{1.0}} & \cellcolor{yellow}\textbf{\meanCI{23.6}{8.1}} & \cellcolor{yellow}\textbf{\meanCI{21.7}{6.7}} & \meanCI{51.9}{9.7} & \meanCI{24.2}{11.8} \\
  & GEPA         & \meanCI{45.0}{2.8} & \meanCI{19.3}{0.2} & \meanCI{20.4}{0.0} & \cellcolor{yellow}\textbf{\meanCI{61.5}{19.3}} & \meanCI{20.2}{2.9} \\
\hhline{------!{\vrule width 1.2pt}-}

\multirow{5}{*}{\texttt{gemma3-4b}} 
  & Baseline      & 27.9 & 9.3 & 11.4 & 36.5 & 13.5 \\
  & RandomSearch  & \meanCI{35.6}{8.0} & \meanCI{8.4}{5.1} & \meanCI{14.0}{4.7} & \meanCI{67.6}{27.3} & \meanCI{15.4}{4.7} \\
  & MIPROv2       & \meanCI{33.9}{10.2} & \meanCI{8.2}{6.2} & \meanCI{16.5}{13.8} & \cellcolor{yellow}\textbf{\meanCI{71.2}{18.2}} & \meanCI{15.4}{7.3} \\
  & SIMBA         & \cellcolor{yellow}\textbf{\meanCI{37.2}{3.3}} & \cellcolor{yellow}\textbf{\meanCI{20.0}{6.9}} & \cellcolor{yellow}\textbf{\meanCI{19.7}{3.8}} & \meanCI{61.9}{13.2} & \meanCI{19.2}{0.0} \\
  & GEPA          & \meanCI{35.1}{9.0} & \meanCI{9.9}{2.3} & \meanCI{15.4}{13.6} & \meanCI{63.5}{32.6} & \cellcolor{yellow}\textbf{\meanCI{21.8}{13.0}} \\
\hhline{------!{\vrule width 1.2pt}-}

\multirow{5}{*}{\texttt{medgemma-4b-it}} 
  & Baseline      & \cellcolor{yellow}\textbf{57.0} & \cellcolor{yellow}\textbf{26.1} & \cellcolor{yellow}\textbf{37.5} \textcolor{red}{\ding{72}} & 1.9 & 7.7 \\
  & RandomSearch  & \meanCI{53.1}{5.2} & \meanCI{21.3}{18.0} & \meanCI{32.7}{3.0} & \meanCI{27.6}{72.4} & \meanCI{19.6}{4.3} \\
  & MIPROv2       & \meanCI{54.2}{13.0} & \meanCI{21.6}{13.1} & \meanCI{31.1}{13.9} & \meanCI{38.5}{12.0} & \meanCI{20.2}{15.5} \\
  & SIMBA         & \meanCI{54.3}{7.5} & \meanCI{24.8}{11.7} & \meanCI{26.6}{27.2} & \meanCI{55.8}{13.8} & \cellcolor{yellow}\textbf{\meanCI{29.2}{13.6}} \\
  & GEPA          & \meanCI{51.0}{6.7} & \meanCI{22.6}{4.7} & \meanCI{25.7}{17.8} & \cellcolor{yellow}\textbf{\meanCI{65.7}{15.8}} & \meanCI{26.3}{1.5} \\
\hhline{------!{\vrule width 1.2pt}-}

\multirow{5}{*}{\texttt{Qwen2.5-VL-7B}} 
  & Baseline      & 47.0 & \cellcolor{yellow}\textbf{24.3} & 17.2 & 61.5 & 9.6 \\
  & RandomSearch  & \meanCI{45.9}{2.8} & \meanCI{21.4}{17.8} & \cellcolor{yellow}\textbf{\meanCI{23.0}{7.6}} & \meanCI{82.7}{12.1} & \meanCI{13.8}{7.0} \\
  & MIPROv2       & \meanCI{46.2}{2.0} & \meanCI{22.0}{10.2} & \meanCI{22.5}{4.2} & \meanCI{79.5}{8.5} & \meanCI{12.8}{5.7} \\
  & SIMBA         & \meanCI{45.5}{1.0} & \meanCI{21.1}{6.7} & \meanCI{22.7}{13.3} & \cellcolor{yellow}\textbf{\meanCI{86.9}{5.8}} & \meanCI{23.4}{9.7} \\
  & GEPA          & \cellcolor{yellow}\textbf{\meanCI{47.4}{4.8}} & \meanCI{21.4}{3.6} & \meanCI{21.3}{2.5} & \meanCI{82.7}{16.7} & \cellcolor{yellow}\textbf{\meanCI{25.0}{8.4}} \\
\hhline{------!{\vrule width 1.2pt}-}

\multirow{5}{*}{\texttt{llama3.2-11B}} 
  & Baseline      & 29.3 & 16.3 & 21.0 & 7.7 & 13.5 \\
  & RandomSearch  & \meanCI{40.8}{8.5} & \meanCI{16.9}{17.1} & \meanCI{18.8}{8.2} & \meanCI{36.5}{64.7} & \meanCI{26.6}{3.2} \\
  & MIPROv2       & \meanCI{39.7}{7.4} & \meanCI{18.1}{15.4} & \meanCI{20.4}{0.0} & \meanCI{49.7}{1.7} & \meanCI{25.3}{8.5} \\
  & SIMBA         & \meanCI{40.9}{7.6} & \cellcolor{yellow}\textbf{\meanCI{22.8}{11.6}} & \meanCI{20.4}{0.0} & \cellcolor{yellow}\textbf{\meanCI{61.5}{42.0}} & \cellcolor{yellow}\textbf{\meanCI{27.6}{1.7}} \\
  & GEPA          & \cellcolor{yellow}\textbf{\meanCI{43.2}{4.6}} & \meanCI{21.9}{4.2} & \cellcolor{yellow}\textbf{\meanCI{21.9}{7.5}} & \meanCI{56.4}{23.1} & \meanCI{25.3}{13.0} \\
\hhline{------!{\vrule width 1.2pt}-}

\multirow{5}{*}{\texttt{gemma3-12b}} 
  & Baseline      & 41.9 & 11.0 & 10.6 & 36.5 & 12.5 \\
  & RandomSearch  & \meanCI{42.2}{2.0} & \meanCI{15.6}{16.2} & \meanCI{19.5}{2.3} & \meanCI{71.8}{40.0} & \meanCI{17.6}{1.7} \\
  & MIPROv2       & \meanCI{41.6}{3.0} & \meanCI{11.7}{9.7} & \meanCI{21.3}{6.4} & \meanCI{70.5}{28.0} & \meanCI{17.3}{0.0} \\
  & SIMBA         & \meanCI{41.9}{5.0} & \cellcolor{yellow}\textbf{\meanCI{25.6}{4.3}} & \cellcolor{yellow}\textbf{\meanCI{22.4}{13.0}} & \meanCI{71.5}{14.2} & \cellcolor{yellow}\textbf{\meanCI{27.2}{13.1}} \\
  & GEPA          & \cellcolor{yellow}\textbf{\meanCI{44.4}{6.8}} & \meanCI{20.4}{9.6} & \meanCI{19.9}{3.8} & \cellcolor{yellow}\textbf{\meanCI{79.2}{4.3}} & \meanCI{23.1}{13.8} \\
\hhline{------!{\vrule width 1.2pt}-}

\multirow{5}{*}{\texttt{gemma3-27b}} 
  & Baseline      & 48.1 & 11.2 & 10.2 & 68.3 & 12.5 \\
  & RandomSearch  & \meanCI{47.8}{0.7} & \meanCI{10.6}{16.3} & \meanCI{22.0}{10.1} & \meanCI{80.8}{24.6} & \meanCI{25.3}{1.7} \\
  & MIPROv2       & \meanCI{47.8}{0.7} & \meanCI{11.4}{6.8} & \cellcolor{yellow}\textbf{\meanCI{22.2}{11.9}} & \meanCI{83.0}{20.0} & \meanCI{23.7}{6.3} \\
  & SIMBA         & \meanCI{48.1}{6.3} & \cellcolor{yellow}\textbf{\meanCI{25.8}{5.8}} & \meanCI{21.8}{15.3} & \cellcolor{yellow}\textbf{\meanCI{86.2}{3.2}} & \cellcolor{yellow}\textbf{\meanCI{30.4}{8.2}} \\
  & GEPA          & \cellcolor{yellow}\textbf{\meanCI{48.8}{3.8}} & \meanCI{16.4}{12.8} & \meanCI{17.9}{6.5} & \meanCI{78.8}{5.6} & \meanCI{24.1}{14.7} \\
\hhline{------!{\vrule width 1.2pt}-}

\multirow{5}{*}{\texttt{Qwen2.5-VL-32B}} 
  & Baseline      & 47.9 & 22.9 & 13.6 & 0.0 & 20.2 \\
  & RandomSearch  & \cellcolor{yellow}\textbf{\meanCI{55.5}{1.0}} & \meanCI{22.6}{9.2} & \meanCI{16.2}{6.0} & \meanCI{0.0}{0.0} & \meanCI{27.0}{5.1} \\
  & MIPROv2       & \meanCI{53.3}{2.0} & \meanCI{24.1}{11.6} & \meanCI{17.1}{14.2} & \meanCI{9.0}{42.5} & \meanCI{22.8}{8.5} \\
  & SIMBA         & \meanCI{54.3}{4.5} & \meanCI{27.8}{14.8} & \meanCI{20.1}{19.5} & \cellcolor{yellow}\textbf{\meanCI{83.2}{6.9}} & \cellcolor{yellow}\textbf{\meanCI{29.5}{7.0}} \\
  & GEPA          & \meanCI{48.1}{1.6} & \cellcolor{yellow}\textbf{\meanCI{28.6}{4.8}} & \cellcolor{yellow}\textbf{\meanCI{21.4}{8.2}} & \meanCI{0.0}{0.0} & \meanCI{25.6}{8.0} \\
\hhline{------!{\vrule width 1.2pt}-}

\multirow{5}{*}{\texttt{Qwen2.5-VL-72B}} 
  & Baseline      & 58.8 & 18.0 & \cellcolor{yellow}\textbf{24.8} & 69.2 & 14.4 \\
  & RandomSearch  & \meanCI{57.1}{6.4} & \meanCI{22.1}{17.5} & \meanCI{22.2}{8.1} & \meanCI{78.5}{1.5} & \meanCI{18.6}{10.6} \\
  & MIPROv2       & \cellcolor{yellow}\textbf{\meanCI{59.1}{0.7}} \textcolor{red}{\ding{72}} & \meanCI{20.2}{17.5} & \meanCI{21.9}{1.0} & \meanCI{79.8}{7.4} & \meanCI{18.0}{11.5} \\
  & SIMBA         & \meanCI{58.2}{5.2} & \cellcolor{yellow}\textbf{\meanCI{30.1}{22.5}} \textcolor{red}{\ding{72}} & \meanCI{20.6}{9.5} & \cellcolor{yellow}\textbf{\meanCI{89.1}{6.5}} \textcolor{red}{\ding{72}} & \cellcolor{yellow}\textbf{\meanCI{30.5}{11.5}} \\
  & GEPA          & \meanCI{57.6}{7.8} & \meanCI{25.4}{4.9} & \meanCI{19.1}{5.1} & \meanCI{86.2}{12.7} & \meanCI{28.4}{14.8} \\
\hhline{------!{\vrule width 1.2pt}-}

\multirow{5}{*}{\texttt{llama3.2-90B}} 
  & Baseline      & 30.8 & 20.8 & 18.8 & 15.4 & 16.3 \\
  & RandomSearch  & \meanCI{43.1}{6.7} & \meanCI{18.6}{9.2} & \meanCI{20.4}{0.0} & \meanCI{53.2}{16.0} & \meanCI{19.6}{20.8} \\
  & MIPROv2       & \meanCI{41.9}{1.8} & \meanCI{19.2}{4.7} & \meanCI{20.4}{0.0} & \meanCI{50.0}{0.0} & \meanCI{22.8}{13.1} \\
  & SIMBA         & \cellcolor{yellow}\textbf{\meanCI{43.6}{5.2}} & \meanCI{21.5}{7.3} & \cellcolor{yellow}\textbf{\meanCI{21.6}{6.0}} & \cellcolor{yellow}\textbf{\meanCI{87.4}{17.9}} & \cellcolor{yellow}\textbf{\meanCI{33.0}{15.3}} \textcolor{red}{\ding{72}} \\
  & GEPA          & \meanCI{40.3}{8.2} & \cellcolor{yellow}\textbf{\meanCI{27.0}{6.7}} & \meanCI{20.4}{0.0} & \meanCI{85.9}{8.0} & \meanCI{29.5}{18.0} \\
\noalign{\vskip 2pt}
\hhline{------!{\vrule width 1.2pt}-}
\noalign{\vskip -1.2pt}
\hhline{------!{\vrule width 1.2pt}-}
\noalign{\vskip 2pt}

\textbf{Average across models} 
  & Baseline     & 41.8 & 16.8 & 18.2 & 31.3 & 12.9 \\
  & RandomSearch & \meanCI{43.2}{0.5} & \meanCI{15.7}{5.3} & \meanCI{20.7}{2.1} & \meanCI{48.6}{9.3} & \meanCI{18.4}{0.7} \\
  & MIPROv2      & \meanCI{44.0}{2.6} & \meanCI{14.7}{5.2} & \meanCI{22.5}{3.1} & \meanCI{54.2}{4.5} & \meanCI{20.6}{3.2} \\
  & SIMBA        & \cellcolor{yellow}\textbf{\meanCI{45.6}{1.3}} & \cellcolor{yellow}\textbf{\meanCI{22.8}{2.9}} & \cellcolor{yellow}\textbf{\meanCI{22.7}{4.3}} & \cellcolor{yellow}\textbf{\meanCI{56.5}{4.0}} & \cellcolor{yellow}\textbf{\meanCI{24.2}{2.7}} \\
  & GEPA         & \meanCI{43.7}{1.4} & \meanCI{17.3}{0.5} & \meanCI{20.5}{3.3} & \meanCI{63.6}{1.3} & \meanCI{22.8}{1.7} \\
\hhline{======!{\vrule width 1.2pt}=}
\end{tabular}
}
\caption{Average performance across three trials spanning five tasks and ten open-source VLMs, comparing a baseline run and four prompt optimization techniques. Yellow cells highlight the best value within each model-task combination while red stars (\textcolor{red}{\ding{72}}) denote the single best score achieved per task. Entries are reported as mean $\pm$ 95\% CI over the three trials.}
\end{table}



Across all models and tasks, the prompt optimization techniques improved performance over the baseline prompts in 45 of 50 model-task combinations.

Among optimizers, SIMBA shows the strongest overall performance, achieving the highest relative improvements in 56\% of benchmark–model combinations. Across its 27 nonzero-baseline wins, the median relative improvement is 73\% (mean 119\%). SIMBA consistently delivers the largest gains across model families. For example, a small general-purpose model (Qwen2.5-VL-3B) improves from an F1 of 7.8 to 23.60 on GastroVision (+203\%), while the optimized medgemma-4b-it, a domain-specialized model, increases its exact-match accuracy on skin-tone classification (DDI) from 1.9\% to 65.70\% (+3358\%), transforming performance from near-random to clinically meaningful. Even larger-scale models benefit substantially: llama3.2-11B rises from 7.7 to 61.53 (+699\%), and llama3.2-90B rises from 15.4 to 87.37 (+467\%) on the same task. GEPA achieves top relative-improvement performance in 26\% of cases, with a median relative improvement of 47\% (mean 319\%). While less frequently the best optimizer, GEPA delivers long-tailed gains, such as boosting medgemma-4b-it on DDI skin-tone classification from 1.9\% to 65.70\% (+3358\%). The highest DDI skin-tone score overall, 89.10, comes from Qwen2.5-VL-72B optimized with SIMBA. MIPROv2 ranks next, leading in 10\% of cases, with a median relative improvement of 95\% (mean 81.1\%). Notably, it produces the highest VQA-RAD score, 59.07, with Qwen2.5-VL-72B. BootstrapFewShotWithRandomSearch leads in 8\% of cases, though its gains are generally smaller (median 2.7\%, mean 6.6\%). For example, Qwen2.5-VL-7B on CheXpert improves from 17.2 to 23.03 (+34\%).

SIMBA achieves the best overall scores on GastroVision, with an optimized Qwen2.5-VL-72B producing the top F1 score of 30.13, and on DDI Diagnosis with llama3.2-90B reaching the highest Top-3 Inclusion rate of 32.97. SIMBA also delivers the strongest performance on DDI Skin Tone, with an exact-match accuracy of 89.10 using Qwen2.5-VL-72B. By contrast, MIPROv2 produces the highest QuasiEM score, reaching 59.07 on VQA-RAD with Qwen2.5-VL-72B. On CheXpert, the highest F1 score remains with the baseline—37.5 for medgemma-4b-it—while the best baseline score overall is 24.8 for Qwen2.5-VL-72B.

Across all evaluations, there are three cases where the best optimized program matches but does not exceed the baseline. Only one of the fifty experiments (llama3.2-11B on CheXpert) shows the baseline outperforming all optimizers. In this instance, optimized prompts either exceed the model’s context window, causing truncation that removes part of the input, or fail to provide sufficiently beneficial instructions. Larger models avoid this issue due to wider context windows, while smaller models still benefit from retained examples even after truncation, leaving this as the sole case where optimization reduces performance.

Additionally, on smaller test sets such as CheXpert and DDI Skin Tone \& Disease, multiple bootstrap trials often converge to similar optimized prompts and hence produce identical predictions and corresponding evaluation scores, resulting in a standard deviation of 0.

\section{Discussion}\label{sec12}

Real-world medical workflows are inherently image-driven and require accurate and reliable visual decision-making. As healthcare moves toward multi-agent, multimodal AI systems, robust performance on image-based tasks becomes essential. Yet current VLM baselines fall short, and existing approaches for improving performance remain data-, compute-, and time-intensive. In contrast, automatic prompt optimization has already transformed LLM performance across diverse tasks, but its potential in multimodal domains remains underexplored. We extend this paradigm to VLMs by building prompting pipelines for complex medical imaging tasks, and demonstrate how automated prompt optimization techniques consistently elevate model performance beyond baselines, particularly without modifying model weights.


By automating prompt tuning for each downstream task and dataset, these techniques provide a scalable way to customize models for the diverse challenges of medical vision-language tasks. SIMBA emerged as the most effective optimizer, delivering the highest relative improvements in 68\% of model-task evaluations. Its reflective process integrates both successes and errors to refine prompts, helping models handle the complex instructions common in clinical scenarios. Although both SIMBA and MIPROv2 combine instruction tuning with in-context learning, SIMBA performs better because its iterative updates incorporate immediate feedback, allowing it to converge on prompts that capture the subtle diagnostic features critical for medical visual reasoning. Notably, GEPA achieved the second-highest performance relying solely on instruction tuning, demonstrating the value of reflective search instruction optimization for vision-language tasks where labeled examples are limited or large sets of images cannot be accommodated within the context window. Additionally, each optimizer incurs a compute cost overhead relative to the baseline, with MIPROv2 at roughly 1×, SIMBA at 1.5×, Random Search at 2.5×, and GEPA at 3× the baseline runtime.

Additionally, the optimization techniques showed performance gains across different model architectures: Qwen models (up to 278\% relative improvement on DDI Skin Tone), Gemma models (up to 143\%), and Llama models (up to 699\%). Within the Qwen series, DDI Skin Tone performance for Qwen2.5-VL-32B rose from 0 to 83.23 with optimization. For cases with nonzero baselines, the largest relative gain is approximately 278\% on DDI Skin Tone (Qwen2.5-VL-3B). In the Gemma family, the greatest relative improvement is about 143\% on DDI Diagnosis (gemma3-27B). The Llama series saw the largest increase, with DDI Skin Tone exact-match improving by nearly 700\% (llama3.2-11B).

In some cases, the optimizers enabled smaller models to outperform their larger counterparts within the same series. For example, the optimized llama3.2-11B reached 61.53\% exact-match on DDI Skin Tone, far exceeding the baseline performance of the much larger llama3.2-90B at 15.4\%. At the same time, even such large models have room for improvement with grounded prompt optimization, with llama3.2-90B improving its DDI Skin Tone exact-match by 467\% over baseline. Even specialized medical models like medgemma-4b-it benefited notably, achieving a 3358\% relative improvement on DDI Skin Tone, highlighting the benefits of automated prompt optimization techniques even for highly domain-specific finetuned models.

While automated prompt optimization offers broad improvements, there remain important limitations to its effectiveness across tasks. In the CheXpert task, few-shot examples often contributed little to performance, and the large image data per input frequently exceeded the context window, preventing optimizers from effectively applying strategies like instruction tuning or iterative feedback. These findings highlight how the practical limits of model context size and the representation of the image data can cap the benefits of prompt optimization for some medical imaging tasks.

The capability to locally host optimized open-source VLM programs is particularly advantageous for medical applications, where data privacy frequently necessitates local secure deployments, and demonstrates the benefits of automated prompt optimization in high-performing medical AI more adaptable. Future research could investigate more complex multimodal compound AI systems, such as those that simultaneously interpret medical scans and clinical notes to generate comprehensive diagnostic reports or systems that integrate radiology images and patient history to assist in clinical decision-making, by optimizing both visual and textual prompts together. 

A key gap in medical vision-language modeling is the lack of systematic evaluations of automated prompt optimization techniques, particularly for vision-based tasks. Our work addresses this by applying structured prompt optimization techniques to a suite of challenging medical benchmarks, demonstrating how to define medical tasks requiring VLM capabilities as composable AI systems, which can then be optimized using automated prompt optimizers. Unlike prior work that relies on manual prompt design or fine-tuning, we show that existing declarative optimizers alone can yield robust, scalable gains in model accuracy without modifying model weights across diverse medical tasks and model architectures. This work highlights the value of prompt optimization for VLMs as a lightweight yet powerful tool to advance medical AI and calls for broader adoption of these techniques in future vision-language research.

\section{Conclusion}\label{sec13}

This work demonstrates that automated prompt optimization can significantly improve the performance of vision-language models on challenging medical tasks without requiring model retraining or access to proprietary data. By systematically applying structured optimization techniques across a diverse set of benchmarks and models, we highlight the effectiveness of inference-time adaptation for domain-specific applications. Our results demonstrate that prompt optimization scales across open-source VLMs and enables smaller models to achieve the capabilities of their larger counterparts, further highlighting the applicability of these techniques for advancing downstream medical AI tasks involving visual data.

\subsection*{Data availability statement}
All datasets analyzed in this study are publicly available. Specifically:
\begin{itemize}
    \item VQA-RAD is available at \url{https://huggingface.co/datasets/flaviagiammarino/vqa-rad}
    \item GastroVision is available at \url{https://github.com/DebeshJha/GastroVision}
    \item CheXpert is available at \url{https://aimi.stanford.edu/datasets/chexpert-chest-x-rays}
    \item DDI is available at \url{https://ddi-dataset.github.io}
\end{itemize}

\noindent
No proprietary or restricted datasets are used.

\backmatter

\begin{appendices}

\section{}\label{secA1}

Sample optimized prompts for tasks (optimized parts are presented in gray blocks)

\medskip
\noindent\textbf{VQA-RAD [MIPROv2 for Qwen2.5-VL-72B]}  

\begin{tcolorbox}[colback=gray!5, colframe=gray!40, boxrule=0.5pt, arc=2pt, breakable]
Answer the question about the image. The answer should be a single word or sentence.  

Analyze the brain MRI image and determine if the gray-white matter junction is altered.  

Analyze the CT scan and determine if there are any abnormalities present. Provide a concise answer based on your findings.  

If no abnormalities are detected, state ``No abnormalities detected''.
\end{tcolorbox}


\noindent\textbf{DDI Skin Tone [GEPA for Qwen2.5-VL-72B]}

\begin{tcolorbox}[colback=gray!5, colframe=gray!40, boxrule=0.5pt, arc=2pt, breakable]
To accurately identify a patient's Fitzpatrick skin tone group from an image, follow these detailed steps:

1. \textbf{Understand the Fitzpatrick Skin Tone Classification:}
   - \textbf{Type I:} Always burns, never tans (pale white; blond or red hair; blue eyes; freckles). This type has very light skin that is highly sensitive to the sun.
   - \textbf{Type II:} Usually burns, tans minimally (white; fair; red or blond hair; blue, hazel, or green eyes). This type has light skin that burns easily but can tan slightly.
   - \textbf{Type III:} Sometimes mild burn, tans uniformly (cream white; fair with golden undertone; dark blonde hair; blue, green, or brown eyes). This type has a light to medium skin tone that tans well but can still burn.
   - \textbf{Type IV:} Burns minimally, always tans well (beige; typical Mediterranean skin tone; brown hair; brown eyes). This type has a medium skin tone that tans easily and rarely burns.
   - \textbf{Type V:} Very rarely burns, tans very easily (brown; typical Middle Eastern skin tones; dark brown hair; brown eyes). This type has a darker skin tone that tans very easily and almost never burns.
   - \textbf{Type VI:} Never burns, tans very deeply (black; very dark brown; black hair; dark brown eyes). This type has the darkest skin tone that never burns and tans very deeply.

2. \textbf{Analyze the Image:}
   - \textbf{Color and Undertones:} Observe the overall color of the skin. Determine if it is very light, light, medium, dark, or very dark. Note any undertones such as golden, beige, reddish, etc. For example, a light skin with a golden undertone might suggest Type III.
   - \textbf{Freckles and Moles:} Note the presence of freckles, moles, or any other pigmentation. Freckles are more common in lighter skin types (Types I and II). Moles can be present in all skin types but are more noticeable in lighter skin.
   - \textbf{Signs of Sun Damage:} Look for signs of aging, discoloration, or other sun-related damage. Sun damage is more visible in lighter skin types. For example, if there are signs of sun damage on a light skin, it might suggest Type I or II.
   - \textbf{Hair and Eye Color:} Note the hair and eye color as they can provide additional clues. For example, dark hair and eyes are more common in darker skin types (Types V and VI). Light hair and eyes are more common in lighter skin types (Types I and II).

3. \textbf{Match the Image to the Fitzpatrick Scale:}
   Use the observations from step 2 to match the skin in the image to the closest Fitzpatrick skin type. Consider the skin's natural color, its reaction to the sun, and any other relevant features. Pay attention to the undertones and the presence of freckles or sun damage. For example, if the skin is light with freckles and signs of sun damage, it is likely to be Type I or II.

4. \textbf{Select the Correct Choice:}
   The choices provided will be in the format 'XY', where 'X' and 'Y' represent two consecutive Fitzpatrick skin types. For example, '12' represents types I and II, '34' represents types III and IV, and '56' represents types V and VI.
   Based on your analysis, select the choice that best matches the skin type you have identified. If the skin appears to be a medium to dark brown color with a significant amount of melanin and no visible signs of freckles or light complexion, it is more likely to be types V or VI.

5. \textbf{Requirements:}
   - Always use the provided list of skin tone groups.
   - Ensure your reasoning is based on the Fitzpatrick skin tone classification.
   - Provide a clear and concise explanation for your choice.
   - Consider the undertones, presence of freckles, and hair and eye color when making your decision.
\end{tcolorbox}


\noindent\textbf{DDI Diagnosis [SIMBA for llama3.2-90B]}

\begin{tcolorbox}[colback=gray!5, colframe=gray!40, boxrule=0.5pt, arc=2pt, breakable]
If the module receives an image of a small, dark, and round lesion with a relatively well-defined border, then it should consider the size and visual characteristics that suggest a benign condition. The presence of a ruler for scale can indicate the lesion is small, which can still be concerning if it has an asymmetrical border, uneven coloration, or if it has changed over time. However, based on the provided choices and the appearance of the lesion, the module should prioritize benign lesions like seborrheic keratosis in its top 3 diagnoses, along with more concerning diagnoses like melanoma and melanocytic nevi.

If the module receives an image with a skin tone that appears darker with a significant amount of melanin present, then it should focus on the overall skin tone and melanin content, and select the highest number in the Fitzpatrick scale that corresponds to the darkest skin types. Avoid being misled by superficial features like freckles or moles, which can be characteristic of lighter skin types. The reddened areas, if any, should be considered a condition and not part of the natural skin tone. Ensure that the reasoning emphasizes the overall darkness and melanin content, aligning with the appropriate Fitzpatrick scale type.

Advice for predict: If the module receives an image with a light complexion and freckles, and the choices provided are '12', '34', and '56', then it should interpret '12' as the closest match to Fitzpatrick skin types 1 and 2, which represent the lighter end of the spectrum. Avoid selecting '34' or '56' as these do not accurately represent the lighter skin tones observed in the image.
\end{tcolorbox}


\noindent\textbf{CheXpert [SIMBA for gemma3-12B]}

\begin{tcolorbox}[colback=gray!5, colframe=gray!40, boxrule=0.5pt, arc=2pt, breakable]
You are an AI doctor specializing in radiology. You are given the patient’s chest
radiograph and a list of possible diagnosis choices. Select all the correct choice(s),
and give the answer as a short response. If none of the choices is correct, output none.
Do not explain.

When presented with a chest radiograph and a list of choices, systematically evaluate the image for features associated with \textit{each} choice before making a decision. In this case, the image showed increased opacity, but the module immediately jumped to consolidation. Instead, consider if the opacity is more consistent with atelectasis (volume loss, crowding of structures) or other options. Prioritize a comprehensive assessment of all choices before settling on a single diagnosis.

When presented with a chest radiograph and a list of choices, systematically evaluate the image for features associated with \textit{each} choice before making a decision. In this case, the image showed increased opacity, but the module immediately jumped to consolidation. Instead, consider if the opacity is more consistent with atelectasis (volume loss, crowding of structures) or other options. Prioritize a comprehensive assessment of all choices before settling on a single diagnosis.
\end{tcolorbox}


\noindent\textbf{GastroVision [SIMBA for Qwen2.5-VL-72B]}

\begin{tcolorbox}[colback=gray!5, colframe=gray!40, boxrule=0.5pt, arc=2pt, breakable]
You are an AI doctor specializing in Gastroenterology. You are given a frame
from a patient’s endoscopy procedure and a list of possible diagnosis choices.
Select only the single correct choice, and give the answer as a short response.
If none of the choices is correct, output none. Do not explain.

If the module receives an endoscopy procedure image and a list of choices, it should focus on analyzing the overall appearance of the gastrointestinal tract, looking for signs of specific pathologies such as polyps, diverticula, or other abnormalities. If none are present, and the image shows a normal mucosal pattern with visible folds, it should select 'Normal mucosa and vascular pattern in the large bowel' as the diagnosis choice. Avoid making assumptions based on specific anatomical features that are not clearly visible or relevant to the provided choices. In the case of the worse trajectory, the module incorrectly identified 'Angiectasia' based on the presence of dilated blood vessels, but it should have considered the overall normal appearance of the mucosa and vascular pattern, which are key indicators for the correct diagnosis.

If the module receives an endoscopy procedure image and a list of choices that includes 'Colorectal cancer', it should focus on analyzing the image for visible masses with irregular borders and changes in the mucosal pattern. The reasoning should specifically mention observing these features, as they are indicative of a potential malignancy. The diagnosis choice should be 'Colorectal cancer' if these features are present. Avoid focusing on less severe conditions like 'Angiectasia' when there are clear signs of a more serious condition.
\end{tcolorbox}


\end{appendices}

\bibliography{sn-bibliography}

\end{document}